\documentclass[a4paper,journal]{IEEEtran}
\usepackage{amsmath,amsfonts}
\usepackage{algorithmic}
\usepackage{array}
\usepackage[caption=false,font=normalsize,labelfont=sf,textfont=sf]{subfig}
\usepackage{xcolor, graphicx}
\usepackage{cite}
\usepackage{url}
\usepackage{color}
\usepackage{hyperref} 
\usepackage{blindtext}

\usepackage{amssymb}
\usepackage{pifont}
\newcommand{\xmark}{\ding{55}}%
\usepackage{dirtytalk}

\begin{document}

\title{Offline and Distributional Reinforcement Learning for Wireless Communications} 

\author{Eslam Eldeeb and Hirley Alves
	\thanks{This research was supported by the Research Council of Finland (former Academy of Finland) 6G Flagship Programme (Grant Number: 346208).}
	
	\thanks{The authors are with the Centre for Wireless Communications (CWC), University of Oulu, Finland. Email: firstname.lastname@oulu.fi. 
}
}

\maketitle
\maketitle

\begin{abstract}
The rapid growth of heterogeneous and massive wireless connectivity in 6G networks demands intelligent solutions to ensure scalability, reliability, privacy, ultra-low latency, and effective control. Although artificial intelligence (AI) and machine learning (ML) have demonstrated their potential in this domain, traditional online reinforcement learning (RL) and deep RL methods face limitations in real-time wireless networks. For instance, these methods rely on online interaction with the environment, which might be unfeasible, costly, or unsafe. In addition, they cannot handle the inherent uncertainties in real-time wireless applications. We focus on \emph{offline} and \emph{distributional} RL, two advanced RL techniques that can overcome these challenges by training on static datasets and accounting for network uncertainties. We introduce a novel framework that combines offline and distributional RL for wireless communication applications. Through case studies on unmanned aerial vehicle (UAV) trajectory optimization and radio resource management (RRM), we demonstrate that our proposed Conservative Quantile Regression (CQR) algorithm outperforms conventional RL approaches regarding convergence speed and risk management. Finally, we discuss open challenges and potential future directions for applying these techniques in 6G networks, paving the way for safer and more efficient real-time wireless systems.

\end{abstract}

\section{Introduction} \label{sec:intro} 

The transition from the fifth-generation (5G) to the sixth-generation (6G) networks unlocks transformative applications, from connected autonomous vehicles (CAVs) and unmanned aerial vehicles (UAVs) to remote surgeries and smart factories. These advancements demand networks that offer ultra-low latency, high reliability, extensive coverage, and energy efficiency, even for massive scenarios. Artificial intelligence (AI) and machine learning (ML) met these demands and are at the forefront, providing innovative solutions redefining next-generation wireless communication capabilities and enabling a wide range of applications \cite{9144301}.

Learning-based solutions play a vital role in shaping future $6$G networks. Deploying AI in wireless networks enables automation and flexibility in handling enormous amounts of data arriving at the core network, enabling a massive number of users, massive multiple-input multiple-output (MIMO), and dynamic resource allocation. In addition, through distributed solutions, AI's scalability optimizes and controls complex, heterogeneous systems at scale. Eventually, the authors in \cite{9170653} argue that AI-powered end-to-end systems may replace traditional model-based approaches.

\begin{figure*}[t!]
    \centering    \includegraphics[width=2.04\columnwidth,trim={0cm 0 0cm 0},clip]{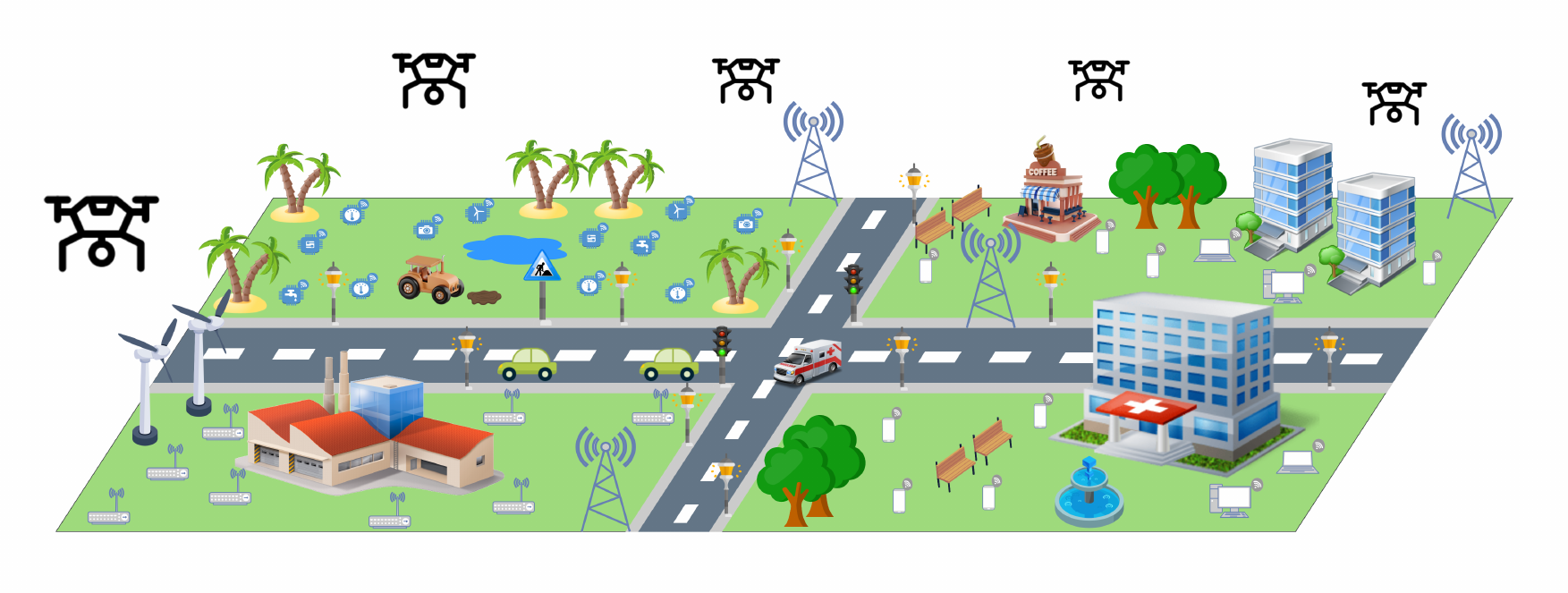} \vspace{2mm}
    \caption{Illustration of a $6$G network with multiple reinforcement learning applications. This includes smart factories, smart agriculture, unmanned aerial vehicle networks, autonomous and connected vehicles, and radio resource management.}
    \vspace{0mm}
    \label{Applications}
\end{figure*}

Reinforcement learning (RL) is crucial for future intelligent 6G networks, offering efficient decision-making in dynamic wireless environments. RL improves decision-making policies through feedback signals received from the wireless environment. 
Deep RL extends this capability by integrating deep neural networks with the RL framework, 
which efficiently optimizes large-scale problems that would otherwise be computationally prohibitive in traditional RL frameworks~\cite{10155733}. RL and deep RL are poised to drive intelligent control and resource management in increasingly complex and massive 6G networks. As shown in Fig.~\ref{Applications}, applications such as smart factories, connected vehicles, resource management, and network slicing will benefit from integrating the RL framework into their deployment.

Building on these capabilities, RL and deep RL have been successfully applied to various wireless communication scenarios. Recent studies demonstrate how these techniques can optimize complex systems in real-world applications. For example, the work in~\cite{9013924} applies deep RL to optimize a UAV's trajectory and scheduling policy, optimizing information freshness in deploying limited-power IoT devices. The authors in~\cite{10714036} propose a fairness-aware RL-based task scheduler in space-air-ground integrated networks.

However, deploying \emph{online RL} in real-life wireless problems presents many challenges \cite{9061001}. A critical issue is that online RL needs continuous interaction with the environment to update and find the optimal policy, which may be impractical due to safety issues or the high cost of data collection \cite{levine2020offl}. Moreover, online RL struggles to identify uncertainties and risks since traditional RL and deep RL algorithms often optimize average performance, maximizing accumulative rewards while neglecting the worst-case scenarios and the inherent uncertainties and risks in the wireless environment \cite{bdr2023}. \emph{Offline RL} and \emph{distributional RL} have emerged as two variants that address these challenges. Offline RL uses a fixed dataset for optimization, eliminating the need for real-time interaction with the environment. Distributional RL shifts the focus to the tail of the distribution of the returns, optimizing the worst scenarios.

This article highlights the advantages of offline and distributional RL as superior alternatives to traditional online RL in wireless communications. Offline RL mitigates online interaction that can be unsafe or costly by performing the optimization using an offline static dataset collected using a behavioral policy, which can be any baseline algorithm. Distributional RL considers the distribution over return, which enables optimizing the tail of the distribution and evaluating sources of risks and uncertainty. We begin with an overview of online and deep RL, highlighting their limitations in real-world wireless networks. Unlink existing literature that either demonstrates traditional online RL or focuses on implementing offline/distributional RL for other domains, \emph{e.g.}, computer science, we introduce a novel framework that combines offline and distributional RL for various wireless applications. We demonstrate the effectiveness of our approach through two critical applications: UAV networks and radio resource management. Our contributions are as follows:
\begin{itemize}
    \item We present a concise overview of the online RL and deep RL, exploring their use in wireless communications and identifying the limitations in dynamic, real-time wireless networks.
    
    \item We propose a novel joint offline and distributional RL framework, enabling offline training while addressing uncertainties in wireless networks.

    \item We validate the framework through simulations on UAV networks and resource management. Finally, we discuss potential challenges, open research problems, and future directions.
\end{itemize}

\section{Overview of Reinforcement Learning} \label{sec:RL}

This section provides an overview of RL and its key components, helping to understand its role in wireless communications. 

\begin{figure*}[t!]
    \centering    \includegraphics[width=2.04\columnwidth,trim={0cm 0 0cm 0},clip]{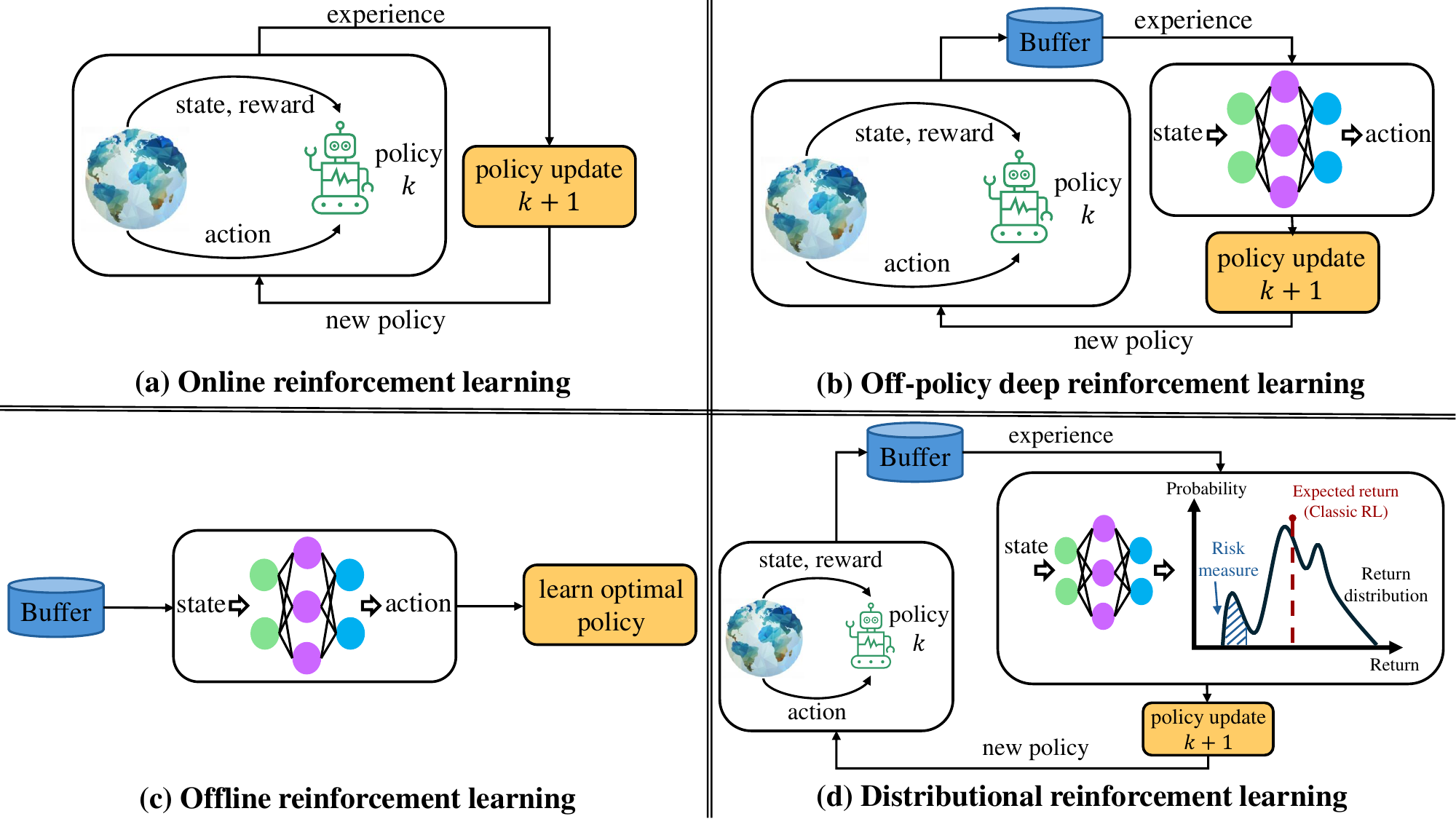} 
    \caption{Reinforcement learning evolution towards offline reinforcement learning: (a) online reinforcement learning, (b) off-policy deep reinforcement learning, (c) offline reinforcement learning, and (d) distributional reinforcement learning. In (a) and (b), the agent can interact with the environment online while optimizing its policy. In (c), the agent can only access a static offline dataset collected previously using behavior policy. In (d), the agent utilizes the return distribution instead of the expected return.}
    \vspace{0mm}
    \label{RL_Variants}
\end{figure*}

\subsection{Reinforcement Learning}
Reinforcement learning (RL), known as the decision-making algorithm, is a branch of AI that enables an agent to learn optimal actions by interacting with the environment. RL environments are typically modeled as Markov decision processes (MDPs). An agent observes its state, takes action, and receives feedback as a reward; this process is illustrated in Fig.~\ref{RL_Variants} (a). A high reward is given if the action benefits the agent; otherwise, a low reward or penalty is applied. The agent aims to maximize its cumulative reward by discovering the optimal policy. 

One key concept in RL is the action-value function (Q-function), mapping state-action pairs to expected rewards. The Q function can be regarded as a reference table that helps the agent evaluate the potential reward of each action in a given state. Therefore, Q-learning is a well-known and widely used model-free RL algorithm that iteratively estimates the optimal Q-function (corresponding consequently to the optimal policy). It calculates the expected reward for a specific action in a particular state, following a certain policy and updating the Q-function with an applicable learning rate.

However, Q-learning has limitations. It requires exploring every possible state-action pair to map out the optimal policy, which becomes infeasible as the environment grows. In real-life wireless applications, networks are heterogeneous, large in dimension, and filled with uncertainties. This "curse of dimensionality" seriously challenges Q-learning's effectiveness, as it struggles to visit every state-action pair in such complex environments.

\subsection{Deep Reinforcement Learning}

To address the limitations of traditional RL in high-dimensional environments, deep reinforcement learning (DRL) introduces deep neural networks as function approximators to estimate the Q-function more efficiently.
To this end, deep RL is a branch of RL that uses deep neural networks as function approximators for estimating Q-functions. As shown in Fig.~\ref{RL_Variants} (b), DRL maps the Q-function by feeding the current state into a neural network, which outputs the best action. The goal is to train the neural network to predict the optimal Q-function by adjusting its weights. One of the most powerful DRL algorithms is the \emph{deep Q-network (DQN)}~\cite{DQNs}, which can solve large-dimension environments in a few iterations. 

DQN is an \emph{off-policy} RL algorithm; that is, it can learn from past experiences stored in a buffer, known as the \emph{replay memory}, rather than interacting with the environment in real time. This allows the agent to use data from different policies to update the current policy more efficiently. 
Over time, several DRL variants have been developed to improve the stability and performance of DQN, including proximal policy optimization (PPO), asynchronous advantage actor-critic (A3C), and deep deterministic policy gradient (DDPG).
%
The use of deep neural networks in these algorithms has led to significant advances in various wireless communication domains, such as optimizing UAV trajectories, network slicing, radio resource management (RRM), CAVs, and power control in cellular networks.

Despite its breakthrough in wireless communication domains, DRL relies on continuous online interaction with the environment to obtain the optimal policy, an obstacle to its scalability in practice. Although off-policy algorithms, such as DRL, can sample from replay memory, they still rely on online interaction with the environment during training. 
%
This presents several challenges: data collection can be expensive and time-consuming in domains like RRM and network slicing, while online training can be unsafe in applications in UAV and CAV networks, where mistakes may lead to catastrophic consequences. In other cases, it would be more efficient to leverage previously collected experiences for training rather than rely solely on the online training process.
%

In the next section, we review recent advances in offline RL and distributional RL, which aim to overcome these challenges by eliminating the need for continuous online interaction.

\section{Offline and Distributional Reinforcement Learning} \label{sec:ODRL}

Offline and distributional reinforcement learning (RL) provide effective alternatives when an online interaction is impractical or unsafe.
In this section, we present the fundamentals of offline and distributional RL and show the benefits of combining them.

\subsection{Offline RL}
Offline RL is a variant of RL and deep RL algorithms that learns optimal policies using only a static dataset, without online interaction with the environment during training~\cite{levine2020offl}. 
Offline RL is applicable when online interaction with the environment is costly or unsafe. As shown in Fig.~\ref{RL_Variants} (c), it optimizes the optimal policy for the agent using a static dataset collected from previous policies, called \emph{behavioral policies}, which are typically sub-optimal and could follow random behaviors, deterministic algorithms, a trained agent online, or even a blend of different policies.

Offline RL is analogous to supervised learning, where a model is trained on a static dataset. 
However, unlike supervised learning, where the goal is to predict accurate outcomes on data similar to the training set, offline RL must generalize to unseen situations, which introduces a \emph{distributional shift}.
%
Thus, offline RL targets achieving high accuracy on data from a different distribution than training data, leading to \emph{out-of-distribution (OOD)} actions. 
%
%
%
This occurs because the state-action pairs seen in the training data often differ from those encountered when deploying the learned policy, leading to overestimating rewards.

%
Recent advances in offline RL address these challenges by adding constraints to the Q-function (or policy). 
These constraints help mitigate the distributional shift by keeping the learned policy close to the behavioral policies, reducing overestimation. 
For example, \emph{Conservative Q-learning (CQL)} \cite{kumar2020conservative}, a well-known offline RL algorithm,  adds a regularization parameter to the Q-learning update (or policy update), known as \emph{conservative penalty parameter}, penalizing OOD actions. 
This ensures that high rewards are not mistakenly assigned to actions the model hasn’t seen. 
%
CQL algorithm can be integrated into classic deep RL architectures, such as DQN, PPO, and actor-critic, by modifying the loss function to include this penalty. 

Other offline RL techniques, such as implicit Q-Learning (IQL), build upon these principles.  
Offline RL is expected to play a critical role in emerging technologies such as digital twins (DTs) and open radio access networks (ORAN), where safe and efficient training is essential.

\subsection{Distributional RL}


The goal of classic RL (online and offline) is to maximize the accumulative \emph{expected return}, making these algorithms inherently \emph{risk-neutral}. They optimize for the average performance, often neglecting the worst-case scenarios.
In contrast, distributional RL overcomes this limitation by considering the distribution of returns, enabling it to address \emph{risk-sensitive} objectives \cite{bdr2023}. 
As shown in Fig.~\ref{RL_Variants} (d), instead of optimizing only for expected returns as in classic RL, 
distributional RL allows optimizing risk-measures, such as \emph{conditional-value-at-risk (CVaR)}. CVaR quantifies the expected return in extreme worst-case scenarios in the environment.

In wireless communication, uncertainties arise due to imperfect knowledge of the environment, stochastic behavior, multiple objectives, and different sources of risks. 
Traditional RL and deep RL struggle with these unpredictable conditions, leading to performance instability. For example, in network slicing and RRM, deep RL algorithms can experience a sudden performance drop due to the environment's stochasticity and perpetual change in system parameters. In contrast, by leveraging the distribution of returns, distributional RL i) offers more stable performance in such imperfect situations and ii) identifies risks and optimizes the network towards risk-free policies.

A key algorithm in distributional RL is the \emph{Quantile regression DQN (QR-DQN)}~\cite{dabney2017distributional}, which estimates the return distribution using quantile regression. 
It approximates the quantile function using a fixed Dirac delta function. The key idea of QR-DQN is to predict not just a single expected reward for an action but the entire range of possible rewards, broken into segments called quantiles. It enables tracking not just the average outcome but also understanding the best-case outcome, worst-case outcome, and outcomes in between, as illustrated in Fig.~\ref{RL_Variants} (d). By doing this, QR-DQN gives a clearer picture of the risks and uncertainties in decision-making, helping to create more reliable and robust strategies, especially in unpredictable environments.
In building a deep distributional RL algorithm such as QR-DQN, the output layer of the neural network must be adjusted to account for the number of actions multiplied by the number of quantile approximations.
%
Recent advances, such as the implicit quantile network (IQN), improve on QR-DQN by
sampling the positions of the Dirac functions instead of keeping them fixed. 

Distributional RL is expected to significantly impact resource management and autonomous vehicle applications, where the ability to handle risks and uncertainties is critical.

\subsection{Offline and Distributional RL}

In real-time wireless problems, offline and distributional RL offer valuable benefits, particularly when combined. Conservative distributional RL merges offline RL techniques, such as \emph{Conservative Q-Learning (CQL)}, with distributional RL algorithms, such as \emph{Quantile Regression DQN (QR-DQN)}, to address the unique challenges of wireless environments. This combination is known as conservative quantile regression (CQR), which stands out because it can be used online and offline. It solves large-dimension problems while identifying risks and uncertainties in dynamic wireless systems. It adapts the conservative principles of CQL to a distributional framework, allowing for more robust optimization in uncertain environments.
Table~\ref{RL_Comparison} compares the evolution of RL algorithms from online RL to more advanced offline and distributional approaches. 

The proposed CQR algorithm effectively enhances adaptability and scalability in dynamic environments by adopting a distributional perspective during policy optimization. The scalability of CQR lies in its ability to efficiently process large-scale systems without sacrificing accuracy or robustness. Its distributional approach enables a detailed understanding of the environment, capturing the full spectrum of potential outcomes rather than merely optimizing for average performance. This ensures that CQR can adapt effectively to uncertainties and unforeseen challenges, maintaining reliability even in complex and rapidly changing systems.

\begin{table}[!t]
\centering
    \caption{A comparison between online RL, off-policy deep RL, offline RL, distributional RL, and offline \& distributional RL.
}
\label{RL_Comparison}
\begin{tabular}{|c|c|c|c|c|}
\hline
 & \textbf{Online} & \textbf{Offline} & \textbf{High dim.} & \textbf{Risk measure}\\ 
\hline
\textbf{Q-learning} & \checkmark & \xmark & \xmark & \xmark\\
\hline                              

\textbf{DQN} & \checkmark & \xmark & \checkmark & \xmark\\
\hline                              

\textbf{CQL} & \checkmark & \checkmark & \checkmark & \xmark\\
\hline                              

\textbf{QR-DQN} & \checkmark & \xmark & \checkmark & \checkmark\\
\hline

\textbf{CQR} & \checkmark & \checkmark & \checkmark & \checkmark\\
\hline

\end{tabular} \vspace{-0mm}
\end{table}

\section{Case Study: Unmanned Aerial Vehicles} \label{sec:UAV_usecase}
This section demonstrates offline and distributional RL use in optimizing the UAVs' trajectory. This experiment considers a UAV serving $10$ limited-power IoT sensors in a $1100$ m $\times$ $1100$ m area. The goal is to jointly minimize the average AoI and the average transmission power by optimizing the UAV's trajectory and scheduling policy. The UAV can serve one device at a time, assuming LoS communication between the UAV and the devices. The AoI of the chosen device resets to $1$, whereas a one-time step increments the AoI of each non-served device. 

We model this problem using MDP. The state space comprises the UAV's position and each device's AoI. In contrast, the action space consists of the movement direction (east, west, north, south, idle) and the chosen device to be served (or being silent without serving any devices). The reward is calculated by summing the individual AoI's weighted sum and transmission power. To introduce uncertainty, we assume an existing risk area in the middle of the grid world, representing an area with poor communication coverage, a high-security restriction area, or a hazardous environment (e.g., a windy area). If the UAV enters this area, there is a small probability (we set it to $10 \%$ in this experiment) that the UAV receives a significant negative reward \cite{eldeeb2024conser}.

\begin{figure}[t!]
    \centering    \includegraphics[width=1\columnwidth,trim={0cm 0 0cm 0},clip]{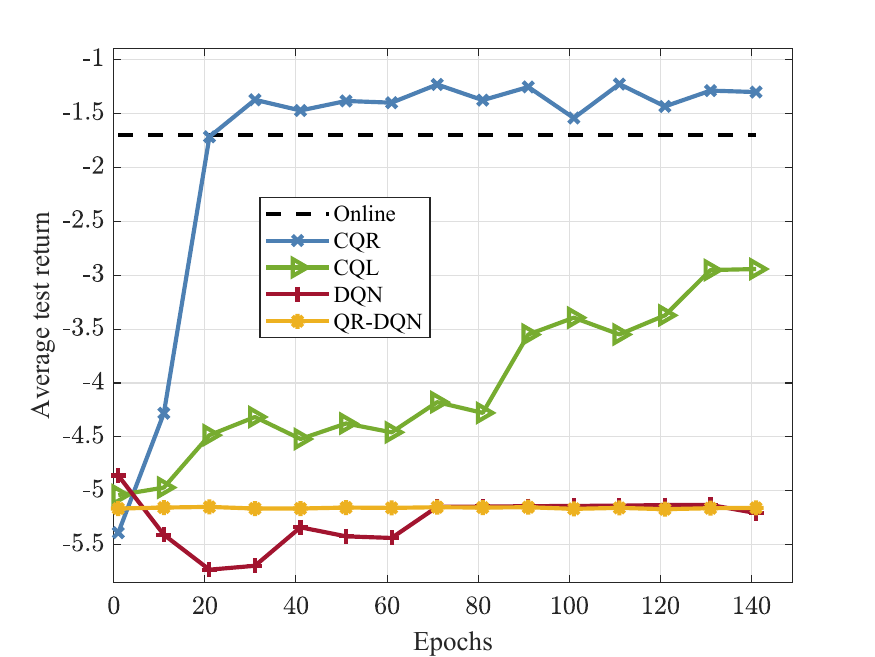} \vspace{2mm}
    \caption{Average test return (normalized by $1000$) over $100$ unique test episodes as a function of the number of training epochs.}
    \vspace{0mm}
    \label{UAV_Fig}
\end{figure}

We compare the proposed CQR algorithm against several baseline RL algorithms trained offline, \emph{i.e.}, DQN, CQL, and QR-DQN. To train these models, we consider collecting an offline dataset using the last $10 \%$ ($3000$ data points) of the experiences (states, actions, next states, rewards) stored in the replay memory of the online DQN agent-trained over $100$ different episodes. To ensure scalability, we validate the algorithm over new unseen $100$ test episodes in testing.

As shown in Fig.~\ref{UAV_Fig}, DQN and QR-DQN fail to converge when directly adapted to an offline setting, highlighting the challenges of applying these algorithms without real-time interaction. CQL shows better convergence but struggles to find the optimum policy due to the uncertainty introduced by the risk area.
%
In contrast, CQR converges to the optimum policy after only $20$ epochs, outperforming the baselines in convergence speed and training efficiency.
Moreover, using a static dataset, CQR outperforms online DQN, showcasing the benefits of combining offline and distributional RL techniques.
%

\begin{figure}[t!]
    \centering    \includegraphics[width=1\columnwidth,trim={0cm 0 0cm 0},clip]{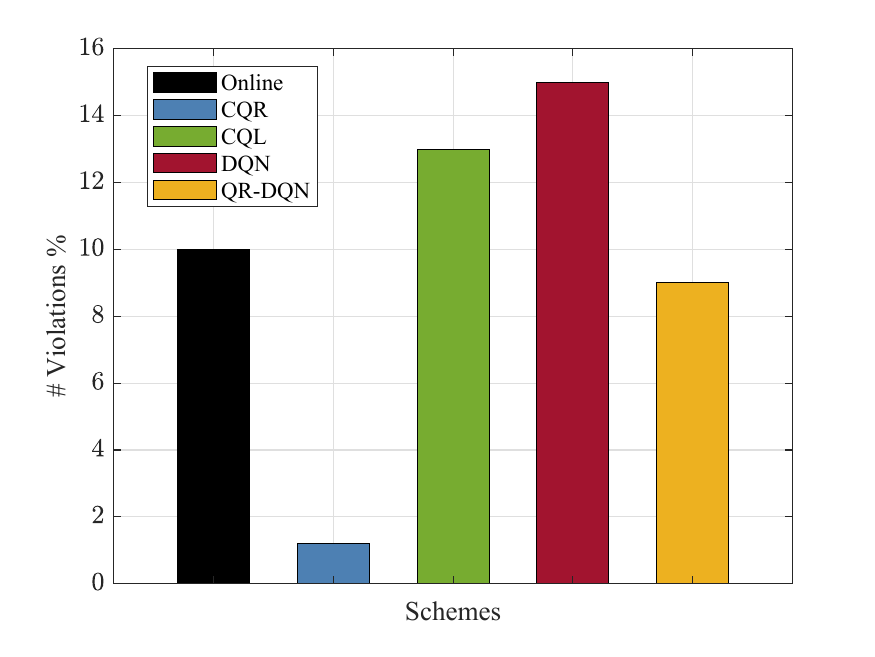} \vspace{2mm}
    \caption{The percentage of violations (the UAV enters the risk region) for different RL schemes averaged over $100$ unique test episodes.}
    \vspace{0mm}
    \label{Violations}
\end{figure}

Next, we illustrate if a model can identify the sources of risk and rare uncertainties in the environment. Thus, in Fig.~\ref{Violations}, we report the percentage of violations, defined as the proportion of time steps the UAV spends in the risk region.
%
We note that online RL has a $10 \%$ violation rate, which equals the probability of receiving a significant negative reward. Both DQN and CQL have high rates of entering the risk region. In contrast, the proposed CQR model has the lowest violation of $1 \%$, demonstrating the benefits of using distributional RL to avoid risk-prone trajectories.

\section{Case Study: Radio Resource Management} \label{sec:RRM_usecase}
In the second case study, we apply offline and distributional RL to RRM in wireless networks. Consider a $100$ m $\times$ $100$ m square area with $4$ randomly deployed access points (APs) and $24$ randomly deployed user equipment (UEs). Each UE moves with a fixed speed $1$ m/s in a random direction, and each AP serves one user at a time. To scale up the problem, we assume that each UE is associated with only one AP during an episode. In addition, UEs associated with each AP are ranked based on their proportional fairness (PF) factor, which indicates how long they have experienced low data rates. 
Each AP can choose one of its top $3$ UEs to serve at a time. 
The objective is to optimize the scheduling policy of the APs that maximizes the rate score (Rscore), defined as the weighted sum of the overall network throughput (e.g., sum rate) and $5$-percentile rate (rate achieved by more than $95 \%$ of the UEs).

We model the RRM problem as an MDP, where the state space comprises the signal-to-interference-plus-noise (SINR) and the PF factors of each AP's top $3$ UEs and the action space comprising the chosen devices for each AP.
%
The reward function is the sum of the weighted PF and the instantaneous rate across all UEs.
Similar to the UAV use case, we collect the offline dataset from the experience of an online DQN agent. 

We compare the CQR algorithm to the baseline models (DQN, QR-DQN, CQL) and  benchmark schemes: \textit{i)} random: each AP chooses its action randomly, \textit{ii)} greedy: each AP chooses the device with the highest PF, \textit{iii)} round robin (RR): each AP serves the users fairly, and \textit{iv)} information-theoretic link scheduling (ITLinQ): it combines full-reuse with TDM, and it achieves a sub-optimal policy \cite{eldeeb2024offlinedistributionalreinforcementlearning}. We consider collecting an offline dataset using the last $10 \%$ ($30000$ data points) of the experiences (states, actions, next states, rewards) stored in the replay memory of the online DQN agent trained over $100$ different episodes. Similar to the previous case study and to ensure scalability, we validate the algorithm over new unseen $100$ test episodes in testing. The states and rewards are normalized to ensure stable training.

\begin{figure}[t!]
    \centering    \includegraphics[width=1\columnwidth,trim={0cm 0 0cm 0},clip]{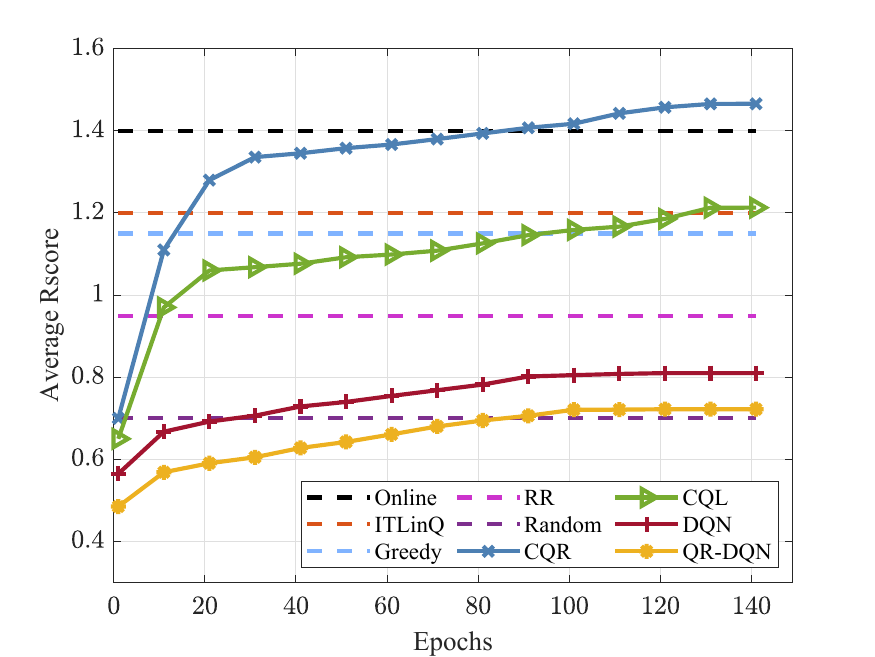} \vspace{2mm}
    \caption{Average test Rscore over $100$ unique test episodes as a function of the number of training epochs.}
    \vspace{0mm}
    \label{RRM_Fig}
\end{figure}

As shown in Fig.~\ref{RRM_Fig}, both DQN and QR-DQN achieve performance similar to the greedy algorithm, surpassing the Rscore of RR and random scheduling. Although CQL outperforms the sub-optimal benchmark ITLinQ, it fails to reach the performance of online RL. In contrast, the proposed CQR is the only algorithm to surpass online RL by leveraging its ability to optimize the whole return distribution rather than focusing solely on the average. 

Our experiments provided several key insights into applying CQR in UAV trajectory optimization and RRM. The dataset quality significantly impacted performance, emphasizing the need for a diverse dataset to prevent overestimation and improve generalization. The choice of loss function was also crucial, as the quantile regression-based loss in CQR enabled more risk-aware decision-making compared to standard RL methods. CQR successfully mitigated risk-prone UAV trajectories and improved fairness in RRM scheduling by optimizing for the worst-case quantiles. These findings demonstrate the potential of offline and distributional RL in wireless networks and suggest future research directions, such as hybrid online-offline training for real-time adaptability.

\section{Open Challenges} \label{sec:Challenges}

Despite their attractive potential deployment in real-world scenarios, offline and distributional RL still face several challenges and future opportunities:

\begin{itemize}

\item \textbf{Offline RL dataset quality:} Offline RL relies heavily on the dataset quality, directly influencing convergence. 
High-quality datasets, rich in diverse and meaningful experiences, lead to faster convergence. In contrast, suboptimal datasets with random or poor experiences can lead to policy divergence or stagnation due to overestimation errors caused by the distributional shift between the behavioral and learned policies.

\vspace{0.6mm}

\item \textbf{Hybrid Online-Offline RL:} Despite its efficiency in convergence only using offline datasets, offline RL can benefit from few online fine-tuning, which helps adapt learned policies to real-time environments \cite{NEURIPS2021_e61eaa38}.
Recently, hybrid online-offline RL algorithms have been proposed to enable periodic online fine-tuning. Hybrid online-offline RL approaches, supported by digital twin technology, allow offline training in simulated environments with limited online exploration in real-world scenarios, improving both safety and efficiency in complex environments like autonomous driving and $6$G networks.

\vspace{0.6mm}

\item \textbf{Distributional RL dimensions:} 
While effective for risk-sensitive optimization, distributional RL faces scalability challenges in high-dimensional environments where the action space explodes. For instance, since distributional RL estimates the return distributions over the actions, the action space, which reflects the size of the output layer of the used neural network, will be the number of quantiles used to estimate the return distribution times the number of actions.
Hence, maintaining efficient convergence is difficult, especially in large-scale wireless systems. Techniques like quantile regression and IQN can alleviate some of these issues, but further work is needed to handle the scale of modern wireless applications such as distributed massive multiple-input multiple output (DmMIMO), reconfigurable intelligent surfaces (RIS), and massive machine-type communication (mMTC). 

\vspace{0.6mm}

\item \textbf{Offline and distributional model-based RL:} 
A major challenge is combining offline and distributional RL with model-based RL due to the difficulty in modeling MDPs and the paradigm shift from predicting future rewards to predicting future states. 
However, this combination can significantly enhance learning efficiency.
%
Nonetheless, errors in model predictions can accumulate, resulting in overestimation of rewards. Adapting conservative algorithms like CQL to model-based approaches could mitigate this by constraining prediction errors, leading to safer and more reliable policies in dynamic environments.
Given the recent advances \cite{kidambi2020morel} and the robustness of the proposed model, it is worthy of future research in combining CQR and model-based RL.

\vspace{0.6mm}

\item \textbf{Offline and distributional multi-agent RL:} CQR can be generalized for multi-agent RL. However, 
challenges in policy coordination and partial observability need further research. In addition, federated learning can be used across multiple datasets collected through various sources in the environment. Thus, it facilitates decentralized learning across multiple agents, allowing collaboration without sharing sensitive data. This is particularly relevant in edge computing and large IoT systems, where communication efficiency and privacy are critical challenges. 

\item \textbf{Fast adaptation with minimal data:} 
An exciting research direction involves combining meta-learning with offline and distributional RL to enable rapid adaptation in new environments using minimal data. Meta-learning, often described as learning to learn, equips agents with the ability to generalize across tasks, extracting transferable knowledge from past experiences. When integrated with offline and distributional RL, this approach could empower agents to quickly adapt to new scenarios without extensive retraining and with small datasets. This direction holds great promise for applications in dynamic and resource-constrained environments.

\end{itemize}

\vspace{2mm}
\section{End Line} \label{sec:discussion}
\vspace{1mm}


Reinforcement learning (RL) has proven to be a powerful tool in wireless communication, offering robust frameworks for control and decision-making. However, deploying traditional online RL in real-time environments presents challenges regarding safety, data collection costs, and handling uncertainties. This article introduced \emph{offline and distributional RL} as viable alternatives, optimizing policies from offline datasets while addressing risks and uncertainties.

Through UAV trajectory optimization and RRM case studies, we demonstrated the effectiveness of the proposed CQR algorithm, which outperforms DQN and CQL in risk management and convergence speed. Beyond these case studies, CQR applies to other wireless problems modeled as MDPs, such as beamforming optimization, interference mitigation, and AR task offloading, due to its ability to perform safe offline training while identifying risks and uncertainties.

Despite promising results, challenges remain, including dataset quality, online fine-tuning, and scalability to multi-agent systems. Addressing these will be critical for advancing RL-driven optimization in next-generation wireless networks.

\vspace{2mm}

\bibliographystyle{IEEEtran}
\bibliography{ref}

\begin{thebibliography}{10}
\providecommand{\url}[1]{#1}
\csname url@samestyle\endcsname
\providecommand{\newblock}{\relax}
\providecommand{\bibinfo}[2]{#2}
\providecommand{\BIBentrySTDinterwordspacing}{\spaceskip=0pt\relax}
\providecommand{\BIBentryALTinterwordstretchfactor}{4}
\providecommand{\BIBentryALTinterwordspacing}{\spaceskip=\fontdimen2\font plus
\BIBentryALTinterwordstretchfactor\fontdimen3\font minus \fontdimen4\font\relax}
\providecommand{\BIBforeignlanguage}[2]{{%
\expandafter\ifx\csname l@#1\endcsname\relax
\typeout{** WARNING: IEEEtran.bst: No hyphenation pattern has been}%
\typeout{** loaded for the language `#1'. Using the pattern for}%
\typeout{** the default language instead.}%
\else
\language=\csname l@#1\endcsname
\fi
#2}}
\providecommand{\BIBdecl}{\relax}
\BIBdecl

\bibitem{9144301}
M.~Z. Chowdhury, M.~Shahjalal, S.~Ahmed, and Y.~M. Jang, ``{6G} wireless communication systems: Applications, requirements, technologies, challenges, and research directions,'' \emph{IEEE Open Journal of the Communications Society}, vol.~1, pp. 957--975, 2020.

\bibitem{9170653}
F.~Tariq, M.~R.~A. Khandaker, K.-K. Wong, M.~A. Imran, M.~Bennis, and M.~Debbah, ``A speculative study on {6G},'' \emph{IEEE Wireless Communications}, vol.~27, no.~4, pp. 118--125, 2020.

\bibitem{10155733}
P.~Cheng, Y.~Chen, M.~Ding, Z.~Chen, S.~Liu, and Y.-P.~P. Chen, ``Deep reinforcement learning for online resource allocation in {IoT} networks: Technology, development, and future challenges,'' \emph{IEEE Communications Magazine}, vol.~61, no.~6, pp. 111--117, 2023.

\bibitem{9013924}
M.~A. Abd-Elmagid, A.~Ferdowsi, H.~S. Dhillon, and W.~Saad, ``Deep reinforcement learning for minimizing age-of-information in {UAV}-assisted networks,'' in \emph{2019 IEEE Global Communications Conference (GLOBECOM)}, 2019, pp. 1--6.

\bibitem{10714036}
G.~Sun, Y.~Wang, H.~Yu, and M.~Guizani, ``Proportional fairness-aware task scheduling in space-air-ground integrated networks,'' \emph{IEEE Transactions on Services Computing}, vol.~17, no.~6, pp. 4125--4137, 2024.

\bibitem{9061001}
N.~Kato, B.~Mao, F.~Tang, Y.~Kawamoto, and J.~Liu, ``Ten challenges in advancing machine learning technologies toward {6G},'' \emph{IEEE Wireless Communications}, vol.~27, no.~3, pp. 96--103, 2020.

\bibitem{levine2020offl}
\BIBentryALTinterwordspacing
S.~Levine, A.~Kumar, G.~Tucker, and J.~Fu, ``Offline reinforcement learning: Tutorial, review, and perspectives on open problems,'' 2020. [Online]. Available: \url{https://arxiv.org/abs/2005.01643}
\BIBentrySTDinterwordspacing

\bibitem{bdr2023}
M.~G. Bellemare, W.~Dabney, and M.~Rowland, \emph{Distributional Reinforcement Learning}.\hskip 1em plus 0.5em minus 0.4em\relax MIT Press, 2023, \url{http://www.distributional-rl.org}.

\bibitem{DQNs}
V.~Mnih, K.~Kavukcuoglu, D.~Silver, A.~Rusu, J.~Veness, M.~Bellemare, A.~Graves, M.~Riedmiller, A.~Fidjeland, G.~Ostrovski, S.~Petersen, C.~Beattie, A.~Sadik, I.~Antonoglou, H.~King, D.~Kumaran, D.~Wierstra, S.~Legg, and D.~Hassabis, ``Human-level control through deep reinforcement learning,'' \emph{Nature}, vol. 518, pp. 529--33, 02 2015.

\bibitem{kumar2020conservative}
A.~Kumar, A.~Zhou, G.~Tucker, and S.~Levine, ``{Conservative {Q}-Learning for Offline Reinforcement Learning},'' in \emph{NeurIPS}, vol.~33, 2020, pp. 1179--1191.

\bibitem{dabney2017distributional}
W.~Dabney, M.~Rowland, M.~Bellemare, and R.~Munos, ``Distributional reinforcement learning with quantile regression,'' in \emph{Proceedings of the AAAI Conference on Artificial Intelligence}, vol.~32, no.~1, 2018.

\bibitem{eldeeb2024conser}
E.~Eldeeb, H.~Sifaou, O.~Simeone, M.~Shehab, and H.~Alves, ``Conservative and risk-aware offline multi-agent reinforcement learning,'' \emph{IEEE Transactions on Cognitive Communications and Networking}, pp. 1--1, 2024.

\bibitem{eldeeb2024offlinedistributionalreinforcementlearning}
\BIBentryALTinterwordspacing
E.~Eldeeb and H.~Alves, ``Offline and distributional reinforcement learning for radio resource management,'' 2024. [Online]. Available: \url{https://arxiv.org/abs/2409.16764}
\BIBentrySTDinterwordspacing

\bibitem{NEURIPS2021_e61eaa38}
\BIBentryALTinterwordspacing
T.~Xie, N.~Jiang, H.~Wang, C.~Xiong, and Y.~Bai, ``Policy finetuning: Bridging sample-efficient offline and online reinforcement learning,'' in \emph{Advances in Neural Information Processing Systems}, vol.~34.\hskip 1em plus 0.5em minus 0.4em\relax Curran Associates, Inc., 2021, pp. 27\,395--27\,407. [Online]. Available: \url{https://proceedings.neurips.cc/paper_files/paper/2021/file/e61eaa38aed621dd776d0e67cfeee366-Paper.pdf}
\BIBentrySTDinterwordspacing

\bibitem{kidambi2020morel}
\BIBentryALTinterwordspacing
R.~Kidambi, A.~Rajeswaran, P.~Netrapalli, and T.~Joachims, ``Morel: Model-based offline reinforcement learning,'' in \emph{Advances in Neural Information Processing Systems}, vol.~33.\hskip 1em plus 0.5em minus 0.4em\relax Curran Associates, Inc., 2020, pp. 21\,810--21\,823. [Online]. Available: \url{https://proceedings.neurips.cc/paper_files/paper/2020/file/f7efa4f864ae9b88d43527f4b14f750f-Paper.pdf}
\BIBentrySTDinterwordspacing

\end{thebibliography}

\vspace{2mm}

\section*{Biographies}
\footnotesize

\vspace{2mm}

\noindent\textbf{ESLAM ELDEEB} received a B.Sc. degree in electrical engineering from Alexandria University, Egypt, in 2019 and an M.Sc. degree from the University of Oulu, Finland, in 2021, where he is currently pursuing a Ph.D. degree with the Centre for Wireless Communication. He is actively working on massive connectivity and ultrareliable low-latency communication. His research interests are machine-type communication and machine learning for wireless communication networks. \\

\noindent \textbf{HIRLEY ALVES} (S’11–M’15) is an Associate Professor on Machine-type Wireless Communications at the Centre for Wireless Communications, University of Oulu, where he leads Massive Wireless Automation Theme in the 6G Flagship Program. He earned his B.Sc. and M.Sc. in electrical engineering from the Federal University of Technology-Paraná, Brazil, 2010, 2011, respectively, and holds a dual D.Sc. from the University of Oulu and UTFPR, 2015. His research interests are massive and critical MTC, satellite IoT, distributed processing and learning, and B5G and 6G technologies.  


\end{document}